\documentclass[11pt]{article}

\usepackage[preprint]{acl}

\usepackage{times}
\usepackage{latexsym}

\usepackage[T1]{fontenc}

\usepackage[utf8]{inputenc}

\usepackage{microtype}
\usepackage{inconsolata}
\usepackage{graphicx}

\usepackage{amsmath,amssymb}
\usepackage{array}
\usepackage{booktabs}
\usepackage{algorithm}
\usepackage{algorithmic}

\newcommand{\systemname}{TactiNavi}
\newcommand{\parahead}[1]{\noindent\textbf{#1}\hspace{0.5em}}

\title{TactiNavi: Serving and Revising an LLM Game Coach with Explicit Strategy Conditions}

\author{
  \textbf{Qi Liu\thanks{Equal contribution.}\textsuperscript{1}},
  \textbf{Xiaoyang Yuan\footnotemark[1]\textsuperscript{1,2,3}},
  \textbf{Zhuomeng Zhang\textsuperscript{1}},
  \textbf{Yubin Ruan\textsuperscript{1}}, \\
  \textbf{Xinyi Mou\textsuperscript{1}},
  \textbf{Wenjin Wang\textsuperscript{1}},
  \textbf{Hanying Jiao\textsuperscript{1}},
  \textbf{Di Wu\textsuperscript{1}}, \\
  \textbf{Jinke Li\textsuperscript{1}},
  \textbf{Jingxuan Hong\textsuperscript{1}},
  \textbf{Ke Feng\textsuperscript{1}},
  \textbf{Zixun Sun\textsuperscript{1}} \\[6pt]
  \textsuperscript{1}Tencent \quad
  \textsuperscript{2}Tongji University \quad
  \textsuperscript{3}Shanghai Innovation Institute
}
\hypersetup{
  pdftitle={TactiNavi: Serving and Revising an LLM Game Coach with Explicit Strategy Conditions},
  pdfauthor={Qi Liu, Xiaoyang Yuan, Zhuomeng Zhang, Yubin Ruan, Xinyi Mou, Wenjin Wang, Hanying Jiao, Di Wu, Jinke Li, Jingxuan Hong, Ke Feng, Zixun Sun}
}

\begin{document}

\maketitle

\begin{abstract}
A deployed LLM coaching agent answers requests whose correct response depends on player state and game version, so advice that is right for one player can be wrong for the next. We describe \systemname{}, the serving system behind the LLM coach of a commercial open-world game, which makes strategy applicability an explicit artifact shared by offline compilation, online execution, failure diagnosis, and release. An offline compiler turns versioned game evidence and expert guidance into strategy packages with explicit prerequisites, the serving plane evaluates those prerequisites as three-valued conditions over structured player state and records decision traces, and a control plane uses the resulting condition--rule dependencies to validate revisions on held-out traffic before release. In a retrospective evaluation of 1,122 logged coaching turns from three deployment periods, condition-bearing packages raise correctness from 57.9\% to 82.0\% over atomic facts and reduce strategy misapplication under changed player state from 18.1\% to 10.4\%, while trace-based claim verification reaches 86.4\% precision at a median reply latency of 2.07\,s. The deployment also showed that most apparent fixes are generation noise, because an unchanged-system rerun recovers 38.4\% of failing turns and only one of 15 shipped revision units passes the release checks retrospectively. We distill these results into lessons on treating applicability as a first-class serving object, sizing traces for later diagnosis, and building the null model into the release path.
\end{abstract}

\section{Introduction}
\label{sec:intro}

Deployed LLM agents increasingly answer requests whose correctness depends on live state rather than on the request alone. A coaching assistant for a strategic game advises on lineups, resource use, and matchups, and the same advice can be right for one player and wrong for another because players own different characters, have made different progress, and play under different game versions. Serving such a system is therefore not only a retrieval problem. The system must decide, per request, whether a piece of strategic advice actually applies, and it must keep making that decision as the game changes underneath the deployment.

Retrieval and grounded synthesis supply the content that such an assistant needs \citep{lewis2020rag,sarthi2024raptor,shao2024storm}, and offline compilation makes that content available at request time. Applicability is a separate problem. When the prerequisites of a strategy stay embedded in prose, the reply model has to notice them, decide whether the current state satisfies them, and explain itself when it cannot tell. A lineup that requires a support role makes the difficulty concrete. The requirement is trivial to check against a roster, yet a prose description leaves the check to generation. Explicit conditions move that work out of generation and into a deterministic evaluation over structured state, where a missing field yields \emph{unknown} instead of a guess.

The same representation also determines what the system can do when a recommendation turns out to be wrong. A player complaint may target an earlier turn, and the disputed claim may rest on flawed strategic knowledge, on a prerequisite that nobody wrote down, on a rule that failed to enforce a valid prerequisite, or on stale player state. Separating these causes requires a record of how the response was produced, and repairing them requires knowing which knowledge and which execution rules depend on each other. A knowledge edit that introduces a new prerequisite is incomplete until some rule can evaluate it, and a rule edit may invalidate the packages that rely on it.

Revision then becomes a measurement problem. A candidate that fixes the reported case may leave the underlying mechanism broken, and a candidate that improves the targeted behavior may damage traffic that nobody thought to check. Generation variability compounds the difficulty, because a rerun of an unchanged system already recovers a substantial share of previously failing interactions. Release decisions therefore need held-out cases drawn from the conditions and rules a revision touches, together with an independent measurement of what the revision actually changed.

We present \systemname{}, a serving system that carries explicit applicability conditions through the whole lifecycle of a coaching agent. Game-specific information enters through three interfaces, namely versioned knowledge and expert guidance, a schema for player state, and tools for game-mechanic calculations. A Strategic Knowledge Compiler synthesizes evidence-grounded strategy packages offline. An Agent Execution Plane evaluates package conditions against player state and game version, recommends only strategies whose prerequisites hold, offers alternatives or asks for missing information otherwise, and records decision traces that link claims to evidence, condition outcomes, and execution rules. A Diagnosis and Repair Evaluator uses those traces to locate the response that feedback targets, verify its claims, and attribute contradictions to the first unsupported stage. An Evolution Control Plane uses the dependency links between conditions and rules to select affected held-out interactions, construct condition-derived counterexamples, and compare reported gains with independently measured effects before a versioned release.

We instantiate \systemname{} as the coaching assistant of a commercial open-world game with strategic battles and evaluate the deployment retrospectively on 1,122 logged coaching turns from three consecutive periods separated by two production revision rounds. Strategy packages reach 82.0\% correctness against 57.9\% for atomic facts, and under controlled state changes they reduce strategy misapplication from 18.1\% to 10.4\%. Claim verification reaches 86.4\% precision on player doubts and localizes doubts that refer back to earlier turns, which preceding-turn baselines cannot do. Median reply generation takes 2.07\,s, with a 95th percentile of 5.48\,s. The revision experiments explain why the control plane is needed. An unchanged system recovers 38.4\% of previously failing turns when simply rerun, only one of 15 shipped revision units satisfies the release criteria, and one unit that gains 25.0 points inside its declared scope loses 20.8 points outside it. Section~\ref{sec:discussion} condenses these measurements into lessons for agents that serve state-dependent advice.

Our contributions are:
\begin{itemize}\setlength{\itemsep}{1pt}\setlength{\parskip}{0pt}\setlength{\topsep}{2pt}
    \item \textbf{Condition-bearing knowledge and an offline compiler.} We separate what a strategy says from when it applies, compile evidence-grounded packages with explicit prerequisites, counterexamples, and instructions, and evaluate those prerequisites as three-valued conditions over structured player state and game version.
    \item \textbf{A serving plane that records enough to repair itself.} Decision traces bind response claims to evidence, condition outcomes, and consuming rules, which supports target-turn localization, claim verification, first-unsupported-stage attribution, and the derivation of dependent knowledge and execution repair targets.
    \item \textbf{Dependency-scoped revision control.} We select held-out cases from the conditions and rules a revision touches, add condition-derived counterexamples, replay candidates against an unchanged-system baseline, and gate release on validated gains and scope-aware regressions.
\end{itemize}

\section{Related Work}
\label{sec:related-work}

\parahead{Knowledge serving for LLM agents.}
Retrieval methods give models access to external evidence \citep{lewis2020rag,guu2020realm,jiang2023flare,asai2024selfrag}, and synthesis methods organize that evidence into reusable artifacts \citep{sarthi2024raptor,shao2024storm}. Agent architectures such as ReAct, CoALA, and Talker-Reasoner structure the loop that consumes such artifacts, combining reasoning, tool calls, and interaction \citep{yao2023react,sumers2024coala,christakopoulou2024talker}. Tool-use work studies how models learn to call external functions and how those calls are planned and executed \citep{schick2023toolformer,qin2024toolllm,kim2024llmcompiler}. State-Grounded Conditioning (SGC) brings structured user state and deterministic rules into the wrappers around a user-facing agent \citep{liu2026sgc}. \systemname{} builds on that grounding mechanism and adds a compiled representation of strategy applicability, so that a prerequisite becomes an object that is evaluated, traced, and revised rather than a sentence the model must interpret.

\parahead{Game agents and coaching systems.}
Game-playing agents use documentation, execution feedback, reusable skills, and retrieval to plan and act \citep{wu2023spring,wang2023deps,wang2023voyager,hu2024pokellmon}. Coaching supports decisions that players make themselves. MOBA Coach analyzes match data for tactical review \citep{horst2021mobacoach}, video-based coaching identifies moments worth commenting on \citep{akie2026gamecoach}, ExInCOACH couples reinforcement learning with language models for state-aware tutoring \citep{hua2026exincoach}, and observational studies of esports coaching emphasize situational analysis and actionable explanation \citep{lee2025crafting}. These systems motivate guidance that adapts to circumstances. Our interest is in the serving and maintenance side of that requirement, where the same condition both selects a strategy and scopes a repair.

\parahead{Failure attribution and agent revision.}
Agent improvement spans reflection and self-critique \citep{shinn2023reflexion,madaan2023selfrefine}, tool-grounded correction \citep{gou2024critic}, and prompt or pipeline optimization \citep{pryzant2023protegi,khattab2024dspy}. The Detection-Aware Attribution Framework (DAAF) formulates failures in terms of editable system attributes that are identified from controlled interventions \citep{yuan2026daaf}. Who\&When attributes failures in multi-agent traces in a single pass \citep{zhang2025whowhen}. At the system level, Ouroboros and ModularRSI improve an agent harness through reviewed changes and recurring deficiencies \citep{razzhigaev2026ouroboros,wu2026modularrsi}, and RRSI and GDPevo study how self-improvement can be constrained or evaluated so that gains generalize \citep{xia2026rrsi,zhou2026gdpevo}. \systemname{} combines target-turn localization and claim verification with condition--rule dependency tracing, which lets one attribution step name both the knowledge and the execution rule that a repair has to touch.

\parahead{Release engineering for learned systems.}
Serving learned components safely has a long literature on hidden technical debt and production readiness \citep{sculley2015technicaldebt,breck2017mltestscore}. Recent work evaluates deployed agents against production-aligned signals and audits reusable skills with verifiers \citep{wang2026linkedin,huang2025auditedskillgraph}, while benchmark studies show how easily evaluation integrity degrades when the evaluator is inside the optimization loop \citep{atinafu2026rewardhackingagents,thaman2026rewardhacking}. \systemname{} treats release as a measurement stage with its own held-out traffic, keeps the release scorer outside candidate development, and reports what an unchanged system already achieves so that a candidate gain can be compared against that floor.

\section{Workload and Requirements}
\label{sec:workload}

\parahead{Requests and state.}
The workload consists of player requests that arrive inside a session, each accompanied by dialogue history and a structured snapshot of player state. The snapshot covers owned characters, resources, progress, and the game version that the session runs against. Roughly three quarters of the coaching turns in our trace ask for strategic guidance such as which lineup to field or how to allocate a limited resource, and the remainder ask about mechanics such as damage calculation or skill interaction. Strategic requests are the ones that need a synthesized strategy rather than a fact lookup.

\parahead{Knowledge and revision cadence.}
Two sources change on different clocks. Game knowledge changes with version updates and expert corrections, and the deployed system is revised in rounds that bundle many small changes. In our evaluation window, the system was revised twice over three consecutive two-day periods, producing three versions and 15 revision units that we group by the strategic sub-intent packages they affect. Revisions arrive after the fact. A player complaint can refer to a response several turns earlier, and the engineer who handles it sees the complaint rather than the state that produced the original answer.

\parahead{Interfaces.}
Three interfaces carry everything that is specific to the game. The first supplies versioned basic knowledge and expert guidance, which the compiler consumes offline. The second defines the state schema that the execution plane reads, including the fields that conditions are allowed to reference. The third exposes tools for mechanic calculations such as damage multipliers and skill interactions. Everything else in \systemname{} is game-independent, and this boundary is what makes the design portable to another title.

\parahead{Requirements.}
Four requirements follow from this workload, and Table~\ref{tab:requirements} maps each to the mechanism that addresses it.

\emph{R1, grounded content.} Advice must be traceable to evidence for the version in force, because a plausible recommendation drawn from an outdated description is worse than no recommendation for a player who acts on it.

\emph{R2, state-dependent applicability.} A strategy may be recommended only when its prerequisites hold for the current player, and the system must distinguish a prerequisite that is false from one that cannot be evaluated. Collapsing the second case into either truth value produces confident advice from missing data.

\emph{R3, diagnosability.} When feedback questions a response, the system must locate the response, decide whether the disputed claim is actually wrong, and name the component that should change. Without a record of the evidence and rules behind the response, every complaint turns into a manual reconstruction.

\emph{R4, bounded revision risk.} A revision must be assessed on interactions that its conditions and rules can reach, compared against the variation of an unchanged system, and checked for effects outside its declared scope. Revision proposals are cheap to generate and expensive to ship, so the gate has to be stricter than the proposal process.

\begin{table}[t]
\centering
\footnotesize
\caption{Serving requirements and the mechanisms that address them.}
\label{tab:requirements}
\setlength{\tabcolsep}{3pt}
\begin{tabular}{@{}>{\raggedright\arraybackslash}p{0.20\columnwidth}>{\raggedright\arraybackslash}p{0.72\columnwidth}@{}}
\toprule
\textbf{Req.} & \textbf{Mechanism} \\
\midrule
R1 & Versioned packages with evidence, prerequisites, and applicable versions (Sec.~\ref{sec:compilation}). \\
R2 & Three-valued conditions over state, evaluated before generation (Sec.~\ref{sec:online}). \\
R3 & Traces and condition--rule dependencies for attribution (Sec.~\ref{sec:diagnosis}). \\
R4 & Scoped validation, counterexamples, and release gates (Sec.~\ref{sec:evolution}). \\
\bottomrule
\end{tabular}
\end{table}

\section{System Design}
\label{sec:design}

\subsection{Overview}
\label{sec:overview}

\begin{figure*}[t]
\centering
\includegraphics[width=0.92\textwidth]{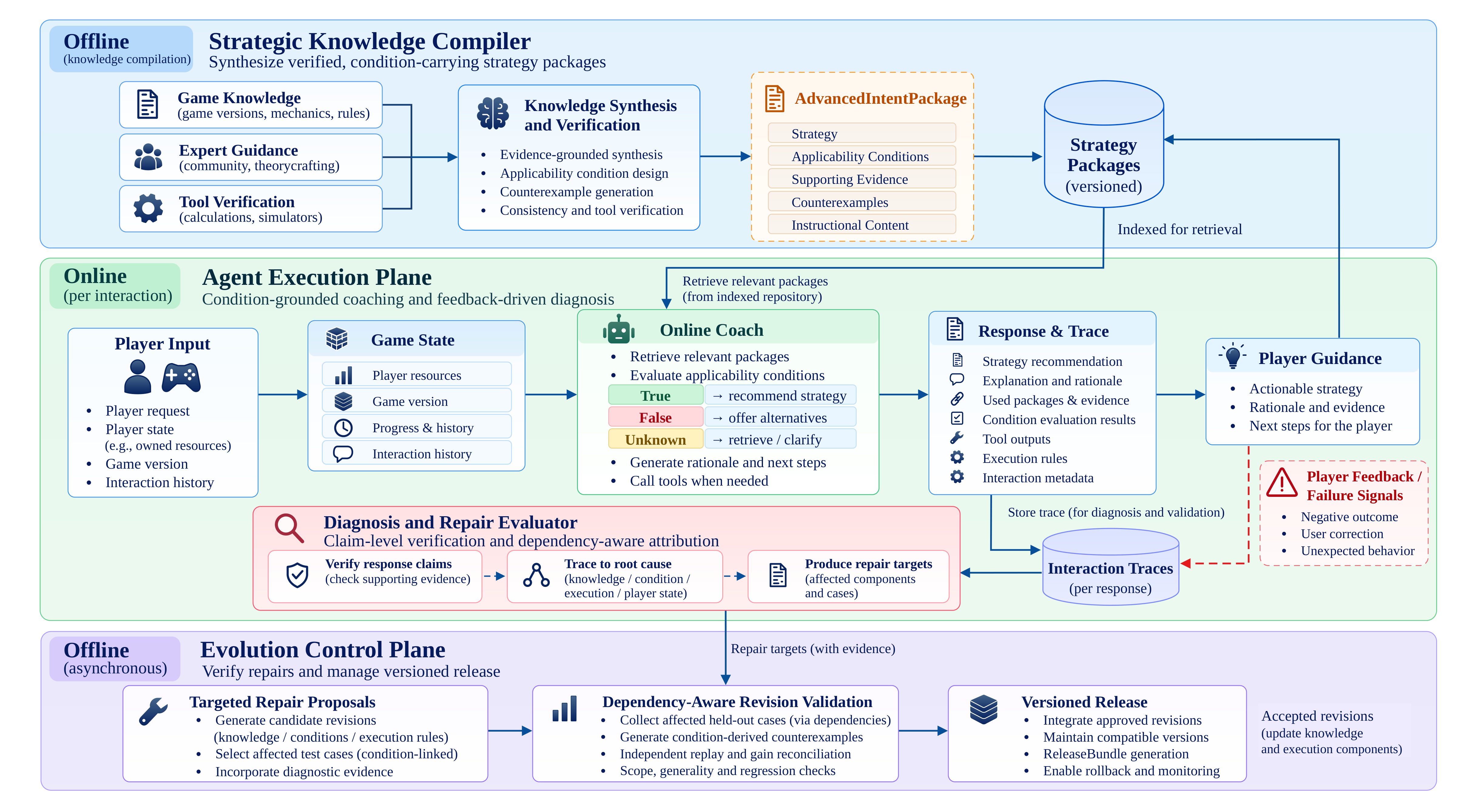}
\caption{Overview of \systemname{}. The Strategic Knowledge Compiler supplies versioned strategy packages for online condition evaluation. The Agent Execution Plane generates guidance and records decision traces, which support feedback-driven diagnosis. Attributed repair targets enter the Evolution Control Plane for independent validation, and accepted revisions update knowledge and execution components for subsequent coaching.}
\label{fig:system}
\end{figure*}

Figure~\ref{fig:system} shows \systemname{} as a compiler and two planes that share one representation, together with an evaluator that connects them. The Strategic Knowledge Compiler runs offline and produces versioned strategy packages. The Agent Execution Plane serves requests, evaluates package conditions against the current snapshot, and records a decision trace for every response. The Diagnosis and Repair Evaluator reads those traces after player feedback arrives and proposes candidate revisions. The Evolution Control Plane validates each candidate on held-out traffic before it enters the serving path. Applicability conditions and the links between conditions and execution rules are the objects that cross all four components, which is why a prerequisite written once can be evaluated online, traced during diagnosis, and used to scope a release decision.

\subsection{Strategic Knowledge Compiler}
\label{sec:compilation}

\parahead{Compilation tasks.}
The compiler builds strategies offline from basic game knowledge and expert guidance. Each task declares a strategic intent, a scope, the versions it applies to, and completion criteria. A knowledge agent gathers evidence from character attributes, skills, counter relations, game rules, and battle statistics. Deterministic tools verify the calculations that the evidence depends on, covering damage multipliers and skill interactions, and a critic tests assumptions and searches for counterexamples. Every resulting claim keeps its evidence, its prerequisites, and the versions under which it holds.

\parahead{Package structure.}
Verified claims become versioned strategy packages that carry applicability conditions, counterexamples, and instructional explanations. A condition states a requirement over player state or game version and evaluates to \emph{true}, \emph{false}, or \emph{unknown} when the available information is insufficient to decide. A lineup strategy may require a support role, for instance, and leave the choice of which eligible owned character fills that role to online execution. Instructions explain why a strategy works, how to apply it, and when to abandon it. Consistency checks and expert review qualify a package as a candidate for versioned release.

\parahead{Dependency index.}
Each condition is linked to the state fields it reads and to the execution-plane rules that evaluate it. These links turn a knowledge artifact into a unit of change with a known blast radius. A requirement that existing rules can already evaluate is updated inside its package, while a new prerequisite that no rule can evaluate also requires a rule change. The same links, combined with decision traces, identify the recorded interactions that a proposed knowledge or execution change would reach.

\subsection{Agent Execution Plane}
\label{sec:online}

\parahead{State-grounded conditioning.}
Building on SGC \citep{liu2026sgc}, the execution plane, which we call the Coach, conditions replies on intent, state, and interaction history through Perception, Grounding, and Interaction wrappers. The state interface supplies player observations, owned resources, progress, and game version. Information that the interface cannot supply stays \emph{unknown} and is never filled in by the reply model.

\parahead{Request path.}
The Coach first classifies the request into an intent and sub-intent, then retrieves the packages that match the player's goal, sub-intent, and game version. The Grounding wrapper checks ownership and resources, and conditional knowledge rules evaluate the remaining prerequisites while deriving the facts that execution needs, such as which owned characters satisfy a required support role. The three-valued outcomes then select a response mode. Satisfied conditions permit the strategy, violated conditions produce an alternative that explains the unmet requirement, and unknown conditions trigger retrieval or a clarification question. Bounded game-tool calls resolve remaining character comparisons, damage estimates, and matchup constraints, with independent calls issued in parallel when possible \citep{kim2024llmcompiler}. The resulting decisions and instructions condition the reply model, which writes the response.

\parahead{A worked request.}
A player asks which lineup to field. Retrieval returns a package whose condition requires a support-role character, and the grounding rule that checks ownership finds no eligible character in the roster, so the condition evaluates to \emph{false} and the Coach answers with an alternative that names the missing role instead of recommending the package. Had the roster been absent from the snapshot, the outcome would be \emph{unknown} and the reply would ask which characters the player owns. Both paths record the same fields, which is what a later complaint about this turn is diagnosed against.

\parahead{Decision traces.}
Each interaction produces a trace that links the response to the selected packages and component versions, the player state that was read, the condition identifiers and their outcomes, the supporting evidence, the tool results, and the reply context. Traces are written on the serving path, so their cost has to stay small relative to generation, yet the same fields are what make the system replayable. Controlled replay holds requests, histories, player states, planning decisions, and tool outputs fixed while varying only knowledge and reply-stage rules, which turns a production interaction into a controlled comparison without re-running the rest of the pipeline.

\subsection{Diagnosis and Repair Evaluator}
\label{sec:diagnosis}

The evaluator follows the DAAF formulation of failures as editable system attributes \citep{yuan2026daaf}. For coaching, it combines target-turn localization, claim verification, and knowledge--execution dependencies to name repair targets for an earlier response or for a single claim inside it. It can also conclude that no repair is needed, or leave attribution unresolved when the evidence does not decide the question.

\parahead{Detection and localization.}
Player doubts, corrections, frustration, tool failures, and empty retrieval results trigger review, together with a sample of unflagged interactions. The evaluator locates the response that the feedback targets, retrieves its trace, and judges the original recommendation against the evidence that was available when it was produced. Judging an old response against current knowledge would attribute every knowledge update to the response rather than to the knowledge.

\parahead{Claim verification and evidence chains.}
The evaluator decomposes the targeted response into atomic claims labeled supported, contradicted, or unknown with respect to versioned knowledge, player resources, and recorded tool outputs. For a contradicted claim it first asks whether the reply misread the context it was given. If the context itself was wrong, it walks the evidence chain through context assembly, routing, tool invocation, and knowledge retrieval to find the first stage that lacks support. Condition identifiers then connect the disputed claim to its prerequisites and to the rules that evaluate them. A flawed prerequisite implicates the package, a valid prerequisite that execution failed to enforce implicates a rule, and inaccurate resources implicate state processing. A verdict record captures the evidence and the implicated attributes, including cases with several targets or none.

\parahead{Repair effect.}
An attributed candidate $\delta$ replaces an editable attribute or a dependent set of knowledge and execution attributes. The evaluator replays the current configuration $S$ and the candidate $S^{\delta}$ on matched development cases $D_{\mathrm{sel}}$, holding requests, player states, game versions, and tool results fixed. With $Q_i(S)$ denoting the correctness score of case $i$, averaged over repeated generations under a fixed claim-level rubric and executable checks, the paired effect is
\begin{equation}
\label{eq:repair-effect}
\widehat{\Delta}_{\mathrm{sel}}(\delta)
:= \frac{1}{|D_{\mathrm{sel}}|}
\sum_{i\in D_{\mathrm{sel}}}
\bigl[Q_i(S^{\delta})-Q_i(S)\bigr].
\end{equation}
A gain on $D_{\mathrm{sel}}$ is therefore treated as a claim that the control plane has to verify.

\subsection{Evolution Control Plane}
\label{sec:evolution}

The control plane assesses revisions on held-out cases with a fixed release protocol that is separate from candidate development. It uses the same correctness definition as repair assessment, while validation cases and release records stay independent of candidate selection.

\parahead{Proposals and dependency closure.}
A proposal records the triggering traces, the targeted attributes, the intended correction, the declared scope, the component versions, and the development gain. Knowledge edits are prepared through the compiler, and condition--rule dependencies determine which execution changes are required for the edit to take effect. Dependent edits are evaluated together as one versioned candidate, because a knowledge change without its rule change would ship a prerequisite that nothing enforces.

\parahead{Condition-aware validation.}
The held-out set $D_{\mathrm{val}}$ combines interactions exposed to the affected conditions or rules, condition-derived counterexamples, and cases for broader regression assessment. For a revised support-role requirement, state variants represent players with an eligible owned character, players without one, and players whose ownership is unknown. The variants preserve unrelated context, and rephrased requests test whether a correction generalizes past the wording that triggered it. Satisfied conditions test that valid strategies remain available, violated conditions test that the Coach still guides the player toward a feasible alternative, and unknown conditions test that incomplete information is treated as unresolved. The validation set therefore mirrors the semantics of the revised condition rather than the surface form of the reported case.

\parahead{Gain reconciliation.}
The control plane re-executes $S$ and $S^{\delta}$ on $D_{\mathrm{val}}$ and computes the paired effect $\widehat{\Delta}_{\mathrm{val}}(\delta)$ by applying Eq.~\ref{eq:repair-effect} to the validation cases. The gap between reported and validated gain is
\begin{equation}
\label{eq:reconcile-gap}
\gamma(\delta)=\widehat{\Delta}_{\mathrm{sel}}(\delta)
-\widehat{\Delta}_{\mathrm{val}}(\delta).
\end{equation}
Decision traces mark the cases that $\delta$ can reach, either because modified knowledge enters their reply context or because a modified rule applies to them. Exposed cases split into those inside and outside the declared repair scope, and the rest stay unexposed, so the validated effect decomposes as
\begin{equation}
\label{eq:reconcile-decomp}
\widehat{\Delta}_{\mathrm{val}}(\delta)
=\sum_{g\in\{\mathrm{in},\mathrm{out},0\}}
\pi_g\,\widehat{\Delta}_{g}(\delta),
\end{equation}
where $\pi_g$ is the share of validation cases in group $g$ and $\widehat{\Delta}_g$ is their mean paired effect. The decomposition distinguishes a scoped improvement from a change that redistributes correctness across traffic, and a large positive $\gamma(\delta)$ signals that the development gain did not reproduce on independent cases. The recovery criterion below turns this signal into a release condition.

\parahead{Release gates.}
A candidate is eligible only if the lower confidence bound of its in-scope gain is positive, recovery on exposed validation failures reaches at least half the recovery observed on development failures, effects on unexposed cases stay within unchanged-system rerun variation, and neither out-of-scope correctness nor answer coverage drops by more than three percentage points. Algorithm~\ref{alg:release} states the procedure. Candidates that fail any check stay quarantined with their evidence.

\begin{algorithm}[t]
\caption{Dependency-scoped revision validation}
\label{alg:release}
\begin{algorithmic}[1]
\STATE \textbf{input:} configuration $S$, candidate $\delta$, development cases $D_{\mathrm{sel}}$
\STATE apply $\delta$ and close over dependent conditions and rules
\STATE select $D_{\mathrm{val}}$ from exposed cases, counterexamples, and rephrased requests
\STATE measure $\widehat{\Delta}_{\mathrm{sel}}$ on $D_{\mathrm{sel}}$ by Eq.~\ref{eq:repair-effect}
\STATE measure $\widehat{\Delta}_{\mathrm{val}}$ and its group terms on $D_{\mathrm{val}}$
\STATE \textbf{checks:} in-scope lower bound $>0$, recovery at least half of development recovery, unexposed shift within rerun variation, out-of-scope loss within tolerance
\IF{any check fails}
\STATE \textbf{return} quarantine
\ENDIF
\IF{generality check finds a case-specific change}
\STATE \textbf{return} quarantine or a scoped exception after review
\ENDIF
\STATE release $S^{\delta}$ with rollback record and predecessor version
\end{algorithmic}
\end{algorithm}

\parahead{Generality and rollback.}
The generality check asks whether a repair addresses a reusable condition or mechanism across affected player states and rephrased requests. Corrections tied to specific entities, case identifiers, or the wording of the triggering interaction remain quarantined, while evidence-backed corrections to narrowly applicable mechanics can be accepted as scoped exceptions after expert confirmation. An accepted candidate becomes a release bundle that records its evidence, reported and validated gains, versions, and rollback predecessor. Dependent edits share a release boundary, and a later hard-constraint violation rolls the bundle back to that predecessor \citep{sculley2015technicaldebt,breck2017mltestscore}.

\parahead{Coupled edits.}
When a candidate changes both knowledge $K$ and execution $E$, the control plane measures the combined effect and, when feasible, each part separately. The interaction $I=\widehat{\Delta}_{K+E}-\widehat{\Delta}_{K}-\widehat{\Delta}_{E}$ captures the departure from additive effects on matched cases. A joint candidate has to pass the gates as a compatible unit, and Section~\ref{sec:reconciliation-results} reports a measured interaction of this kind.

\section{Evaluation}
\label{sec:evaluation}

We evaluate \systemname{} on recorded player interactions and on controlled replay. The evaluation asks whether condition-bearing knowledge improves correctness, whether traces support diagnosis of real complaints, and whether the release gate separates durable revisions from changes that only look like improvements.

\subsection{Setup}
\label{sec:setup}

\parahead{Deployment and data.}
We instantiate \systemname{} as the coaching assistant of a commercial open-world game with strategic battles, where players assemble battle lineups from owned characters and ask for guidance on strategies and mechanics. The evaluation uses logged requests, histories, player states, tool outputs, and reply contexts. Player and session identifiers are de-identified before analysis, and the controlled comparisons, condition variants, and planted revisions run as offline replays of frozen traces that never reach players. The serving path and its traces ran in production throughout the window, and the two revision rounds shipped to players. The release checks of Section~\ref{sec:evolution} were not part of those rounds, so we apply them afterwards with criteria fixed in advance and ask how many shipped revisions they would have admitted.

\parahead{Periods and replay.}
Two revision rounds produce three system versions, $S_1$, $S_2$, and $S_3$, over consecutive two-day periods $T_1$, $T_2$, and $T_3$. Both rounds use preceding interactions for development and leave $T_3$ sealed. The first round mainly changes reply-stage rules and the second mainly updates game knowledge, and together they comprise 15 revision units grouped by the strategic sub-intent packages they affect. We sample 82 sessions per period, which yields 1,122 coaching turns, 360 of them in $T_3$. Controlled comparisons replay the target reply while holding requests, histories, player states, planning decisions, and tool outputs fixed, and only knowledge and reply-stage rules follow the configuration under test. Each configuration generates two responses per turn, so that reruns of an unchanged system measure generation variation.

\parahead{Models and measures.}
A self-hosted DeepSeek-V4-Flash model \citep{xu2026deepseek} generates responses and provides judgments over shuffled, configuration-blind outputs against versioned game evidence. A response is \emph{correct} when it contains no contradicted claims and passes executable checks on owned characters, available skills, counter relations, and resource-dependent abilities. It \emph{passes} when it is also instructive and averages at least 1.5 across applicable relevance, explanation (Q2), and actionability (Q3) scores, each ranging from 0 to 2.

\parahead{Annotation and statistics.}
Game-domain experts annotate 294 turns from $T_1$ and $T_2$, reaching $\kappa=0.69$ on correctness. After calibration on $T_2$, the judge agrees with expert correctness labels on 85.7\% of 154 held-out $T_1$ turns when executable checks are included. The rubric is fixed after calibration and applied unchanged across replayed configurations. We report 95\% confidence intervals from session-clustered bootstrap resampling and use exact McNemar tests for stable paired outcomes.

\subsection{Strategic Knowledge and Applicability}
\label{sec:results}

We compare four knowledge configurations on 197 strategic turns from $T_3$ that use strategy packages. All four share the player state, tool outputs, reply-stage rules, and response model, so differences come from the knowledge form alone. \emph{Atomic facts} supplies character and mechanic information. An \emph{unstructured guide} expresses the content of a package and its rule outputs as prose without explicit condition markers. The \emph{strategy package} keeps structured applicability conditions, and the $-$\,\emph{cond.\ rules} ablation removes conditional-rule outputs while retaining the rest of the package.

We also construct 108 matched player-state variants from 36 turns whose strategies depend on a lineup-wide resource requirement, making each requirement true, false, or unknown. Guides keep their original text, while packages re-evaluate the affected condition. A violation occurs when advice applies a strategy under a false condition, withholds it under a true condition, or treats an unknown condition as settled.

\begin{table}[t]
\centering
\scriptsize
\caption{Knowledge configurations on $T_3$ strategic turns. Corr., Pass, and Viol. are percentages. Viol. is measured on false and unknown variants. Contr. counts contradicted claims and violations. Q2 and Q3 range from 0 to 2. Tok. counts knowledge tokens.}
\label{tab:knowledge}
\setlength{\tabcolsep}{2.2pt}
\begin{tabular}{@{}lccccccc@{}}
\toprule
Knowledge & Corr. & Contr. & Q2 & Q3 & Pass & Tok. & Viol. \\
\midrule
Atomic facts & 57.9 & 0.91 & 1.02 & 1.21 & 41.9 & 767 & 10.4 \\
Guide & 82.5 & 0.28 & 1.27 & 1.43 & 62.7 & 1298 & 18.1 \\
Package & 82.0 & 0.28 & 1.32 & 1.44 & 65.2 & 1477 & 10.4 \\
\quad $-$ cond.\ rules & 75.6 & 0.43 & 1.19 & 1.43 & 61.4 & 1246 & 21.5 \\
\bottomrule
\end{tabular}
\end{table}

Table~\ref{tab:knowledge} shows that the package reaches 82.0\% correctness against 57.9\% for atomic facts, a 24.1-point gain with a 95\% CI of [16.4, 31.7] and $p<0.001$, and raises the pass rate from 41.9\% to 65.2\%. Explanation quality rises from 1.02 to 1.32 and actionability from 1.21 to 1.44. On the 75 turns that use conditional-rule outputs, removing those rules lowers correctness from 84.7\% to 69.3\% ($p<0.001$).

The unstructured guide matches the package on recorded turns, at 82.5\%. On the state variants with false or unknown conditions, however, it violates applicability more often than the package, at 18.1\% against 10.4\%, and removing conditional rules pushes violations to 21.5\%. Relative to the guide, the package reduces violations by 7.6 points with a 95\% CI of [2.1, 15.4], and relative to the ablation by 11.1 points with a 95\% CI of [1.9, 22.8]. The guide comparison rests on 72 variant units from 36 base turns, and the exact McNemar test on stable pairs, with 5 discordant pairs favoring the package and none favoring the guide, gives $p=0.062$, so we read the violation reduction as consistent evidence of an effect rather than as a settled one. The gap is widest when the relevant state is unknown, where violations are 26.4\% for the guide and 15.3\% for the package. Atomic facts violate as rarely as the package, at 10.4\%, because they carry no strategy that could be misapplied, and they pay for this with the lowest correctness on recorded turns. The comparison therefore separates two things that a single correctness number conflates. Synthesis improves what the system says, and explicit conditions determine whether it says it when the strategy applies.

\subsection{Failure Diagnosis}
\label{sec:diagnosis-results}

We evaluate 130 player-doubt turns from 39 sessions in $T_1$, including 48 doubts adjudicated by experts, of which 36 are founded, meaning the disputed response was indeed wrong, and 12 unfounded. Precision and recall are computed over founded doubts, and a method rejects an unfounded doubt when it declines to call the response a failure. Baselines treat doubt as failure, judge the feedback directly, route failures into categories, or apply Who\&When trace attribution \citep{zhang2025whowhen}, and we also ablate turn localization. All methods use the same underlying model.

\begin{table}[t]
\centering
\footnotesize
\caption{Failure diagnosis on expert-adjudicated $T_1$ doubts. Prec. and Rec. are percentages for founded doubts. Rej. counts rejected unfounded doubts out of 12. Lag. is delayed-feedback localization (\%).}
\label{tab:diagnosis}
\setlength{\tabcolsep}{3.2pt}
\begin{tabular}{@{}lcccc@{}}
\toprule
Method & Prec. & Rec. & Rej. & Lag. \\
\midrule
Doubt as failure & 75.0 & 100.0 & 0/12 & 0.0 \\
Direct feedback & 75.6 & 94.4 & 1/12 & 0.0 \\
Category routing & 88.0 & 61.1 & 9/12 & 0.0 \\
Who\&When & 77.8 & 97.2 & 2/12 & 0.0 \\
\systemname{} & 86.4 & 52.8 & 9/12 & 26.7 \\
\quad $-$ turn localization & 80.8 & 58.3 & 7/12 & 0.0 \\
\bottomrule
\end{tabular}
\end{table}

Table~\ref{tab:diagnosis} reports 86.4\% precision and 52.8\% recall for \systemname{}, which rejects 9 of 12 unfounded doubts against 1 for direct feedback and 2 for Who\&When. Category routing reaches 88.0\% precision and 61.1\% recall, so the trace-based method gives up a little precision and some recall in exchange for rejecting unfounded complaints and locating the responsible earlier response in 26.7\% of delayed-feedback failures, which preceding-turn methods cannot do at all. Removing target-turn localization drops precision to 80.8\%. Evidence-based screening and delayed-feedback localization are the two capabilities that the traces add.

\subsection{Revision Gains and Release Validation}
\label{sec:reconciliation-results}

We assess the deployed revisions on the sealed $T_3$ period. Replaying $S_1$, $S_2$, and $S_3$ on the same 360 turns shows that the final version does not differ measurably from the first, with a correctness difference of $+0.6$ percentage points between $S_3$ and $S_1$, a 95\% CI of $[-2.2, 3.4]$, and $p=1.0$. A factorial replay on 26 turns affected by jointly developed first-round revisions separates knowledge-only updates, which gain $+7.7$ points, from the knowledge--execution interaction of $-9.6$ points with a 95\% CI of $[-17.9, -2.0]$. The combination performs below the sum of its parts, which is why dependent edits are released as a unit.

We then apply the release checks retrospectively to the 15 deployed revision units with criteria fixed in advance. Recovery, the share of previously failing turns that a configuration answers correctly, is the signal a development team would normally read as evidence of a repair, so we also measure it for an unchanged system. To test whether the checks distinguish case-specific corrections from reusable ones, we build six planted controls across three failure classes, namely recommending unavailable characters, misstating counter relations, and making unsupported turn-order claims. Each class receives an entity-specific correction and a mechanism-based correction of similar length. The candidates are inserted independently into the $S_3$ reply-stage rules and assessed on $T_3$.

\begin{table}[t]
\centering
\footnotesize
\caption{Independent revision assessment on $T_3$. Rec. is recovery among previously failing turns for the unchanged rerun and among failing exposed turns for planted controls (\%). Gain is the paired correctness difference in percentage points, evaluated in scope for planted controls. Spec. counts case-specific revisions, and Eligible counts those passing release checks. Unchanged reruns use $S_1$, while planted controls use $S_3$.}
\label{tab:reconciliation}
\setlength{\tabcolsep}{3.2pt}
\begin{tabular}{@{}lccccc@{}}
\toprule
Revision & Units & Rec. & Gain & Spec. & Eligible \\
\midrule
Unchanged rerun & -- & 38.4 & $+0.6$ & -- & -- \\
\midrule
Deployed units & 15 & -- & -- & -- & 1/15 \\
\midrule
Planted, entity & 3 & 39.1 & $-0.6$ & 3/3 & 0/3 \\
Planted, mechanism & 3 & 40.7 & $+0.7$ & 0/3 & 0/3 \\
\bottomrule
\end{tabular}
\end{table}

Only one deployed unit satisfies the release criteria. Thirteen show no gain beyond generation variation. The remaining unit improves in-scope correctness by 25.0 points while reducing out-of-scope correctness by 20.8 points, a regression that targeted evaluation would have missed and that the scope decomposition in Eq.~\ref{eq:reconcile-decomp} exposes directly.

Table~\ref{tab:reconciliation} makes the noise floor visible. Rerunning the unchanged system recovers 38.4\% of previously failing turns, close to the 39.1\% and 40.7\% recovery of the planted revisions, yet the planted controls produce paired in-scope gains of only $-0.6$ and $+0.7$ points and all six fail gain reconciliation. Recovery on failing turns is therefore not evidence of a repair. The gate requires a positive in-scope gain before release, and that requirement alone removes every planted candidate.

The generality check classifies the three entity-specific revisions as case-specific and the three mechanism-based revisions as general, reaching 85.7\% agreement with expert labels over 21 units at $\kappa=0.56$. Mechanism-based controls are classified as general even though they fail gain reconciliation, which shows that the two checks answer different questions. Gain reconciliation asks whether correctness improved, and generality asks whether the correction extends beyond the cases that triggered it.

\subsection{Cost and Serving Behavior}
\label{sec:cost}

Reply generation on the 360 $T_3$ turns has a median of 2.07\,s and a 95th percentile of 5.48\,s, with a mean of 2.47\,s. The 269 strategy turns are somewhat slower, at 2.42\,s median and 5.60\,s at the 95th percentile, which is consistent with their longer replies. Condition evaluation runs as rule checks over the structured snapshot and issues no model calls, so the measured latency is dominated by generation rather than by the applicability machinery.

The knowledge representation has a price that is visible in context size. A package costs 1,477 knowledge tokens against 767 for atomic facts and 1,298 for the prose guide, so explicit conditions add 14\% over the guide while cutting violations by 7.6 points on state variants. The ablation shows the reverse trade, at 1,246 tokens with violations of 21.5\%, because conditional rules are cheaper to omit than to evaluate.

Replay is what keeps validation affordable. Only the target reply is regenerated, with planning decisions and tool outputs held fixed, so a candidate can be compared against the current system on hundreds of held-out turns without re-executing the full pipeline. The same mechanism gives the release gate its control condition, because an unchanged system can be replayed under identical inputs and its variation used as the null.

\section{Lessons from Deployment}
\label{sec:discussion}

Operating \systemname{} taught us four lessons that we expect to carry over to other LLM agents whose answers depend on live user state and on a knowledge base that changes under them.

\parahead{Lesson 1: Applicability deserves to be an object.}
The comparison between the prose guide and the strategy package carries the central design argument. The two forms are indistinguishable on recorded turns, at 82.5\% and 82.0\%, and they diverge by 7.6 points once the same strategies meet changed player state. Content quality and applicability control are therefore separable engineering goals, and a system that optimizes only the first will look finished while it silently misapplies what it knows. Making conditions explicit costs 14\% more context than prose and buys a fallback path for missing state, a target for diagnosis, and a scope for release.

\parahead{Lesson 2: Traces must be cheap on the serving path and sufficient off it.}
The trace is the only artifact that survives the request, and it carries the evidence, the condition outcomes, and the rules that produced the response. Because it is written on the serving path, it cannot hold everything. Because diagnosis happens later and often about a different turn, it must hold enough to reconstruct why a claim was made. The 26.7\% delayed-feedback localization and the rejection of 9 of 12 unfounded doubts both depend on that record, and neither is available from the conversation alone.

\parahead{Lesson 3: Independence belongs in the system, not in the process.}
An unchanged system recovers 38.4\% of failing turns, which is close to what the planted revisions recover. Any workflow that accepts a candidate because the reported case now passes will ship changes that do nothing, and 13 of our own 15 shipped units fit that description. Logged correctness stayed at 83.1\%, 84.4\%, and 84.4\% across the three periods even though two revision rounds shipped in between (Appendix~\ref{app:version-replay}). The control plane treats an unchanged rerun as the null and requires a positive in-scope gain before release, which is a strict rule and eliminates all six planted controls. The rule is affordable because replay regenerates only the reply, so validating a candidate on hundreds of held-out turns costs one generation per turn.

\parahead{Lesson 4: Coupled knowledge and execution need a shared release boundary.}
Knowledge edits create prerequisites that rules must evaluate, and rules encode assumptions that packages rely on. The measured interaction of $-9.6$ points on jointly developed changes shows that the two parts cannot be optimized or released in isolation. The dependency closure gives the pair a single boundary, which also makes rollback well defined.

\parahead{Limitations.}
We evaluate one commercial game across two revision rounds, so the transfer to games with different mechanics and to longer revision sequences remains open. The state-variant comparison between packages and guides uses 36 base turns, which leaves the violation reduction short of conventional significance. Replay holds player states, planning decisions, and tool outputs fixed, which characterizes revisions to knowledge and reply rules rather than to the whole pipeline. Expert-attributed failures concentrate in reply execution, so attribution performance on other components is not measured here. The automated judge shares the Coach's underlying model, which makes evaluation sensitive to shared judgment errors. The reported latency covers reply generation and not the complete online workflow.

\section{Conclusion}
\label{sec:conclusion}

We described \systemname{}, the serving system behind a deployed LLM game coach, which treats strategy applicability as an explicit artifact and carries it through compilation, serving, diagnosis, and release. Strategy packages with evidence-grounded content and three-valued conditions raise correctness over atomic facts and reduce misapplication when player state changes, while decision traces make late and ambiguous complaints diagnosable and condition--rule dependencies give a revision a scope. The deployment showed that most shipped revisions did not survive independent measurement, and that a release gate which rejects them costs one generation per held-out turn because the rest of the pipeline is held fixed. We expect the same three ingredients, namely explicit applicability, diagnosis-ready traces, and a replayed null model at release, to matter for other agents that serve state-dependent advice from a changing knowledge base.

\label{sec:main-text-end}

\bibliography{references}

@article{yao2023react,
  title={React: Synergizing reasoning and acting in language models},
  author={Yao, Shunyu and Zhao, Jeffrey and Yu, Dian and Du, Nan and Shafran, Izhak and Narasimhan, Karthik and Cao, Yuan},
  journal={arXiv preprint arXiv:2210.03629},
  year={2022}
}

@article{shinn2023reflexion,
  title={Reflexion: Language agents with verbal reinforcement learning},
  author={Shinn, Noah and Cassano, Federico and Gopinath, Ashwin and Narasimhan, Karthik and Yao, Shunyu},
  journal={Advances in neural information processing systems},
  volume={36},
  pages={8634--8652},
  year={2023}
}

@article{lewis2020rag,
  title={Retrieval-augmented generation for knowledge-intensive nlp tasks},
  author = {Lewis, Patrick and Perez, Ethan and Piktus, Aleksandra and Petroni, Fabio and Karpukhin, Vladimir and Goyal, Naman and K{\"u}ttler, Heinrich and Lewis, Mike and Yih, Wen-tau and Rockt{\"a}schel, Tim and Riedel, Sebastian and Kiela, Douwe},
  journal={Advances in neural information processing systems},
  volume={33},
  pages={9459--9474},
  year={2020}
}

@inproceedings{liu2023agentbench,
  title={Agentbench: Evaluating llms as agents},
  author = {Liu, Xiao and Yu, Hao and Zhang, Hanchen and Xu, Yifan and Lei, Xuanyu and Lai, Hanyu and Gu, Yu and Ding, Hangliang and Men, Kaiwen and Yang, Kejuan and Zhang, Shudan and Deng, Xiang and Zeng, Aohan and Du, Zhengxiao and Zhang, Chenhui and Shen, Sheng and Zhang, Tianjun and Su, Yu and Sun, Huan and Huang, Minlie and Dong, Yuxiao and Tang, Jie},
  booktitle={International Conference on Learning Representations},
  pages={52989--53046},
  year={2024}
}

@article{schick2023toolformer,
  title={Toolformer: Language models can teach themselves to use tools},
  author={Schick, Timo and Dwivedi-Yu, Jane and Dess{\`\i}, Roberto and Raileanu, Roberta and Lomeli, Maria and Hambro, Eric and Zettlemoyer, Luke and Cancedda, Nicola and Scialom, Thomas},
  journal={Advances in neural information processing systems},
  volume={36},
  pages={68539--68551},
  year={2023}
}

@article{hu2024pokellmon,
  title={Pokellmon: A human-parity agent for pokemon battles with large language models},
  author={Hu, Sihao and Huang, Tiansheng and Liu, Ling},
  journal={arXiv preprint arXiv:2402.01118},
  year={2024}
}

@inproceedings{sarthi2024raptor,
  title={Raptor: Recursive abstractive processing for tree-organized retrieval},
  author={Sarthi, Parth and Abdullah, Salman and Tuli, Aditi and Khanna, Shubh and Goldie, Anna and Manning, Christopher},
  booktitle={International Conference on Learning Representations},
  pages={32628--32649},
  year={2024}
}

@inproceedings{shao2024storm,
  title={Assisting in writing wikipedia-like articles from scratch with large language models},
  author={Shao, Yijia and Jiang, Yucheng and Kanell, Theodore and Xu, Peter and Khattab, Omar and Lam, Monica},
  booktitle={Proceedings of the 2024 Conference of the North American Chapter of the Association for Computational Linguistics: Human Language Technologies (Volume 1: Long Papers)},
  pages={6252--6278},
  year={2024}
}

@article{ma2024agentboard,
  title={Agentboard: An analytical evaluation board of multi-turn llm agents},
  author={Ma, Chang and Zhang, Junlei and Zhu, Zhihao and Yang, Cheng and Yang, Yujiu and Jin, Yaohui and Lan, Zhenzhong and Kong, Lingpeng and He, Junxian},
  journal={Advances in neural information processing systems},
  volume={37},
  pages={74325--74362},
  year={2024}
}

@article{madaan2023selfrefine,
  title={Self-refine: Iterative refinement with self-feedback},
  author = {Madaan, Aman and Tandon, Niket and Gupta, Prakhar and Hallinan, Skyler and Gao, Luyu and Wiegreffe, Sarah and Alon, Uri and Dziri, Nouha and Prabhumoye, Shrimai and Yang, Yiming and Gupta, Shashank and Majumder, Bodhisattwa Prasad and Hermann, Katherine and Welleck, Sean and Yazdanbakhsh, Amir and Clark, Peter},
  journal={Advances in neural information processing systems},
  volume={36},
  pages={46534--46594},
  year={2023}
}

@inproceedings{gou2024critic,
  title={Critic: Large language models can self-correct with tool-interactive critiquing},
  author = {Gou, Zhibin and Shao, Zhihong and Gong, Yeyun and Shen, Yelong and Yang, Yujiu and Duan, Nan and Chen, Weizhu},
  booktitle={International Conference on Learning Representations},
  pages={57734--57811},
  year={2024}
}

@article{zheng2023llmjudge,
  title={Judging llm-as-a-judge with mt-bench and chatbot arena},
  author = {Zheng, Lianmin and Chiang, Wei-Lin and Sheng, Ying and Zhuang, Siyuan and Wu, Zhanghao and Zhuang, Yonghao and Lin, Zi and Li, Zhuohan and Li, Dacheng and Xing, Eric P. and Zhang, Hao and Gonzalez, Joseph E. and Stoica, Ion},
  journal={Advances in neural information processing systems},
  volume={36},
  pages={46595--46623},
  year={2023}
}

@article{sumers2024coala,
  title={Cognitive architectures for language agents},
  author={Sumers, Theodore R and Yao, Shunyu and Narasimhan, Karthik and Griffiths, Thomas L},
  journal={arXiv preprint arXiv:2309.02427},
  year={2023}
}

@article{wang2023voyager,
  title={Voyager: An open-ended embodied agent with large language models},
  author={Wang, Guanzhi and Xie, Yuqi and Jiang, Yunfan and Mandlekar, Ajay and Xiao, Chaowei and Zhu, Yuke and Fan, Linxi and Anandkumar, Anima},
  journal={arXiv preprint arXiv:2305.16291},
  year={2023}
}

@inproceedings{wang2023deps,
  title={Describe, Explain, Plan and Select: Interactive Planning with Large Language Models Enables Open-world Multitask Agents},
  author = {Wang, Zihao and Cai, Shaofei and Chen, Guanzhou and Liu, Anji and Ma, Xiaojian and Liang, Yitao},
  booktitle={Proceedings of the 37th International Conference on Neural Information Processing Systems. Red Hook: Curran Associates Inc},
  pages={34153--34189},
  year={2024}
}

@article{wu2023spring,
  title={Spring: Studying papers and reasoning to play games},
  author={Wu, Yue and Min, So Yeon and Prabhumoye, Shrimai and Bisk, Yonatan and Salakhutdinov, Russ R and Azaria, Amos and Mitchell, Tom M and Li, Yuanzhi},
  journal={Advances in Neural Information Processing Systems},
  volume={36},
  pages={22383--22687},
  year={2023}
}

@inproceedings{guu2020realm,
  title={Retrieval augmented language model pre-training},
  author={Guu, Kelvin and Lee, Kenton and Tung, Zora and Pasupat, Panupong and Chang, Mingwei},
  booktitle={International conference on machine learning},
  pages={3929--3938},
  year={2020},
  organization={PMLR}
}

@inproceedings{asai2024selfrag,
  title={Self-rag: Learning to retrieve, generate, and critique through self-reflection},
  author={Asai, Akari and Wu, Zeqiu and Wang, Yizhong and Sil, Avi and Hajishirzi, Hannaneh},
  booktitle={International conference on learning representations},
  pages={9112--9141},
  year={2024}
}

@inproceedings{jiang2023flare,
  title={Active retrieval augmented generation},
  author={Jiang, Zhengbao and Xu, Frank F and Gao, Luyu and Sun, Zhiqing and Liu, Qian and Dwivedi-Yu, Jane and Yang, Yiming and Callan, Jamie and Neubig, Graham},
  booktitle={Proceedings of the 2023 conference on empirical methods in natural language processing},
  pages={7969--7992},
  year={2023}
}

@inproceedings{qin2024toolllm,
  title={Toolllm: Facilitating large language models to master 16000+ real-world apis},
  author = {Qin, Yujia and Liang, Shihao and Ye, Yining and Zhu, Kunlun and Yan, Lan and Lu, Yaxi and Lin, Yankai and Cong, Xin and Tang, Xiangru and Qian, Bill and Zhao, Sihan and Hong, Lauren and Tian, Runchu and Xie, Ruobing and Zhou, Jie and Gerstein, Mark and Li, Dahai and Liu, Zhiyuan and Sun, Maosong},
  booktitle={International Conference on Learning Representations},
  pages={9695--9717},
  year={2024}
}

@article{kim2024llmcompiler,
  title={An llm compiler for parallel function calling},
  author={Kim, Sehoon and Moon, Suhong and Tabrizi, Ryan and Lee, Nicholas and Mahoney, Michael W and Keutzer, Kurt and Gholami, Amir},
  journal={arXiv preprint arXiv:2312.04511},
  year={2023}
}

@inproceedings{zhou2024webarena,
  title={Webarena: A realistic web environment for building autonomous agents},
  author = {Zhou, Shuyan and Xu, Frank F. and Zhu, Hao and Zhou, Xuhui and Lo, Robert and Sridhar, Abishek and Cheng, Xianyi and Ou, Tianyue and Bisk, Yonatan and Fried, Daniel and Alon, Uri and Neubig, Graham},
  booktitle={International Conference on Learning Representations},
  pages={15585--15606},
  year={2024}
}

@inproceedings{mialon2024gaia,
  title={Gaia: a benchmark for general ai assistants},
  author={Mialon, Gr{\'e}goire and Fourrier, Cl{\'e}mentine and Wolf, Thomas and LeCun, Yann and Scialom, Thomas},
  booktitle={International Conference on Learning Representations},
  pages={9025--9049},
  year={2024}
}

@inproceedings{khattab2024dspy,
  title={Dspy: Compiling declarative language model calls into state-of-the-art pipelines},
  author = {Khattab, Omar and Singhvi, Arnav and Maheshwari, Paridhi and Zhang, Zhiyuan and Santhanam, Keshav and Vardhamanan, Sri and Haq, Saiful and Sharma, Ashutosh and Joshi, Thomas T. and Moazam, Hanna and Miller, Heather and Zaharia, Matei and Potts, Christopher},
  booktitle={International Conference on Learning Representations},
  pages={54928--54958},
  year={2024}
}

@inproceedings{pryzant2023protegi,
  title={Automatic prompt optimization with ``gradient descent'' and beam search},
  author={Pryzant, Reid and Iter, Dan and Li, Jerry and Lee, Yin and Zhu, Chenguang and Zeng, Michael},
  booktitle={Proceedings of the 2023 conference on empirical methods in natural language processing},
  pages={7957--7968},
  year={2023}
}

@inproceedings{liu2023geval,
  title={G-eval: NLG evaluation using gpt-4 with better human alignment},
  author={Liu, Yang and Iter, Dan and Xu, Yichong and Wang, Shuohang and Xu, Ruochen and Zhu, Chenguang},
  booktitle={Proceedings of the 2023 conference on empirical methods in natural language processing},
  pages={2511--2522},
  year={2023}
}

@article{sculley2015technicaldebt,
  title={Hidden technical debt in machine learning systems},
  author={Sculley, David and Holt, Gary and Golovin, Daniel and Davydov, Eugene and Phillips, Todd and Ebner, Dietmar and Chaudhary, Vinay and Young, Michael and Crespo, Jean-Francois and Dennison, Dan},
  journal={Advances in neural information processing systems},
  volume={28},
  year={2015}
}

@inproceedings{breck2017mltestscore,
  title={The ML test score: A rubric for ML production readiness and technical debt reduction},
  author={Breck, Eric and Cai, Shanqing and Nielsen, Eric and Salib, Michael and Sculley, D},
  booktitle={2017 IEEE international conference on big data (big data)},
  pages={1123--1132},
  year={2017},
  organization={IEEE}
}

@article{razzhigaev2026ouroboros,
  title={Ouroboros: A self-developing frontier coding agent with reviewed core evolution},
  author={Razzhigaev, Anton and Gritsaev, Andrei and Kaznacheev, Andrei and Dragunov, Nikita and Yampolskiy, Roman and Kuznetsov, Andrei},
  journal={arXiv preprint arXiv:2608.08311},
  year={2026}
}

@article{xia2026rrsi,
  title={RRSI: Regularized recursive self-improvement of agent harnesses},
  author = {Xia, Peng and Han, Rujun and Wang, Zifeng and Chen, Yanfei and Zhuang, Yufan and Lee, Yoonho and Huang, Chengsong and Yu, Han and CuiZhu, Zhongying and Ming, Yifei and Yao, Huaxiu and Gokturk, Burak and Pfister, Tomas and Lee, Chen-Yu},
  journal={arXiv preprint arXiv:2609.24972},
  year={2026}
}

@article{wu2026modularrsi,
  title={ModularRSI: Modular and generalizable recursive harness self-improvement},
  author = {Wu, Siwei and Ren, Jincheng and Li, Yizhi and Li, Haau-Sing and Yang, Chengran and Ma, Xudong and Zhang, Yuxuan and Gu, Weicheng and Yang, Jian and Batista-Navarro, Riza and Zhang, Chuanyi and Liu, Xianglong and Zhou, Ming and Dai, Bryan and Lin, Chenghua},
  journal={arXiv preprint arXiv:2609.14857},
  year={2026}
}

@article{zhou2026gdpevo,
  title={GDPevo: Evaluating Agent Self-Evolution on Real Business Tasks},
  author = {Zhou, Leijun and Liu, Zhihao and Qu, Xiang and Liu, Chenxu and Liu, Yifei and Yu, Yanke and Xu, Jingzhe and Wu, Xuejun and Qian, Buyue and Chen, Xi and Zheng, Yaowei and Hu, Junhao},
  journal={arXiv preprint arXiv:2608.03764},
  year={2026}
}

@article{zhang2025whowhen,
  title={Which agent causes task failures and when? on automated failure attribution of llm multi-agent systems},
  author = {Zhang, Shaokun and Yin, Ming and Zhang, Jieyu and Liu, Jiale and Han, Zhiguang and Zhang, Jingyang and Li, Beibin and Wang, Chi and Wang, Huazheng and Chen, Yiran and Wu, Qingyun},
  journal={arXiv preprint arXiv:2505.00212},
  year={2025}
}

@article{yuan2026daaf,
  title={{DAAF}: From Failure Localization to Editable System Assets in {LLM} Agents},
  author={Yuan, Xiaoyang and Liu, Qi and Ruan, Yubin and Mou, Xinyi and Zhang, Zhuomeng and Wang, Wenjin and Jiao, Hanying and Wu, Di and Xu, Mingye and Bin, Yi and others},
  journal={arXiv preprint arXiv:2609.32498},
  year={2026}
}

@article{liu2026sgc,
  title={State-Grounded Conditioning: Wrapping User-Facing {LLM} Agents Where Direction Depends on Live State},
  author={Liu, Qi and Yuan, Xiaoyang and Ruan, Yubin and Zhang, Zhuomeng and Wang, Wenjin and Wu, Di and Xu, Mingye and Mou, Xinyi and Yin, Xingxi and Feng, Ke and others},
  journal={arXiv preprint arXiv:2609.27606},
  year={2026}
}

@inproceedings{akie2026gamecoach,
  title={Game Coaching System Using 3D CNN and LLM},
  author={Akie, Ryogo and Tu, Yunhao and Murase, Kazuyuki and Alnajjar, Fady},
  booktitle={2026 IEEE 2nd International Conference on Consumer Technology (ICCT-Pacific)},
  pages={594--597},
  year={2026},
  organization={IEEE}
}

@inproceedings{horst2021mobacoach,
  title={MOBA Coach: Exploring and analyzing multiplayer online battle arena data},
  author={Horst, Robin and Lanvers, Micha and von Kacsoh, L{\'a}szl{\'o} and D{\"o}rner, Ralf},
  booktitle={International Symposium on visual computing},
  pages={197--209},
  year={2021},
  organization={Springer}
}

@inproceedings{lee2025crafting,
  title={Crafting champions: an observation study of esports coaching processes},
  author={Lee, Hanbyeol and Kleinman, Erica and Kim, Namsub and Park, Sangbeom and Harteveld, Casper and Lee, Byungjoo},
  booktitle={Proceedings of the 2025 CHI Conference on Human Factors in Computing Systems},
  pages={1--20},
  year={2025}
}

@article{christakopoulou2024talker,
  title={Agents thinking fast and slow: A talker-reasoner architecture},
  author={Christakopoulou, Konstantina and Mourad, Shibl and Matari{\'c}, Maja},
  journal={arXiv preprint arXiv:2410.08328},
  year={2024}
}

@article{hua2026exincoach,
  title={ExInCOACH: Strategic exploration meets interactive tutoring for context-aware game onboarding},
  author={Hua, Rui and Huang, Zhaoyu and Lu, Jinhao and Li, Yakun and Zhao, Na},
  journal={Information Fusion},
  pages={104151},
  year={2026},
  publisher={Elsevier}
}

@article{wang2026linkedin,
  title={Self-evolving Agentic Customer Support System at LinkedIn},
  author={Wang, Chih Hui and Tu, Mengdie and Zhang, Qianyun and Wu, Wei and Zhou, Lili and Shen, Mingqi and Wei, Changshuai},
  journal={arXiv preprint arXiv:2608.10224},
  year={2026}
}

@article{huang2025auditedskillgraph,
  title={Audited skill-graph self-improvement for agentic llms via verifiable rewards, experience synthesis, and continual memory},
  author={Huang, Ken and Huang, Jerry},
  journal={arXiv preprint arXiv:2512.23760},
  year={2025}
}

@inproceedings{atinafu2026rewardhackingagents,
  title={RewardHackingAgents: Benchmarking Evaluation Integrity for LLM ML-Engineering Agents},
  author={Atinafu, Yonas and Cohen, Robin},
  booktitle={2026 IEEE 42nd International Conference on Data Engineering Workshops (ICDEW)},
  pages={27--34},
  year={2026},
  organization={IEEE}
}

@article{thaman2026rewardhacking,
  title={Reward hacking benchmark: measuring exploits in LLM agents with tool use},
  author={Thaman, Kunvar},
  journal={arXiv preprint arXiv:2605.02964},
  year={2026}
}

@article{xu2026deepseek,
  title={Deepseek-v4: Towards highly efficient million-token context intelligence},
  author = {{DeepSeek-AI}},
  journal={arXiv preprint arXiv:2606.19348},
  year={2026}
}

\appendix
\section{Component Details}
\label{app:components}

This section describes how applicability conditions support evidence-chain attribution, condition-aware validation, and the assessment of proposed repairs.

\subsection{Evidence-chain attribution}
\label{app:attribution}

Attribution starts from a contradicted claim and follows the recorded evidence chain from the reply toward its sources. Table~\ref{tab:attribution-stages} lists the stages examined and the repair owner of each. The first stage that lacks support identifies the implicated component. A supported claim requires no repair, and missing evidence leaves attribution unresolved for clarification or expert review. Condition identifiers link a disputed claim to its prerequisites and to the rules that evaluate them, which lets related knowledge and execution targets be assessed together.

\begin{table*}[t]
\centering
\footnotesize
\caption{Evidence-chain stages used for failure attribution. The first unsupported stage determines the implicated repair owner.}
\label{tab:attribution-stages}
\setlength{\tabcolsep}{4pt}
\begin{tabular}{@{}>{\raggedright\arraybackslash}p{0.16\textwidth}>{\raggedright\arraybackslash}p{0.52\textwidth}>{\raggedright\arraybackslash}p{0.22\textwidth}@{}}
\toprule
\textbf{Stage} & \textbf{Unsupported when} & \textbf{Repair owner} \\
\midrule
Reply rendering & The response misstates or exceeds its reply context & Reply execution \\
Context assembly & Available relevant evidence is omitted from the reply context & Reply execution \\
Result merging & A correct result is lost when tool outputs are combined & Upstream execution \\
Routing & A required tool or knowledge source is not selected & Upstream execution \\
Tool input & A tool receives incorrect arguments & Upstream execution \\
Tool output & A tool omits information available from its source & Upstream execution \\
Knowledge source & Required knowledge is incorrect, missing, or inaccessible & Knowledge \\
\bottomrule
\end{tabular}
\end{table*}

\subsection{Condition-aware validation}
\label{app:validation}

Validation cases follow the applicability conditions that a revision modifies and the execution rules that consume them. The control plane retrieves held-out interactions involving these dependencies and builds matched counterexamples by changing the relevant player-state fields while preserving the request and unrelated state. For a strategy that requires an eligible role, the variants represent players with an eligible owned character, players without one, and players whose ownership information is unavailable. These states produce \emph{true}, \emph{false}, and \emph{unknown} outcomes, and resource and version requirements are tested across their relevant boundaries. Rephrased requests vary the original wording while preserving intent and constraints.

Expected behavior is established independently of the proposed revision, using versioned game rules and expert-approved strategy conditions. A satisfied condition permits the strategy, a violated condition calls for an alternative that explains the unmet requirement, and an unknown condition requires additional evidence or clarification. Previously successful interactions involving the affected knowledge or rules provide regression cases, so the validation set measures both the targeted correction and its effect on existing valid guidance.

\subsection{Interpreting a failed reconciliation}
\label{app:reconciliation}

The release checks separate distinct sources of apparent improvement. A large positive reconciliation gap means that the reported development gain exceeds the effect reproduced on independent cases. Changes on unexposed cases beyond unchanged-system variation call for inspecting sampling effects, scorer sensitivity, and changes outside the declared scope. A decline in out-of-scope correctness identifies a regression, and reduced answer coverage or increased deferral suggests that a candidate avoids difficult recommendations. These signals guide further examination of quarantined proposals.

\subsection{Generality and scoped exceptions}
\label{app:generality}

The generality check considers how a repair is expressed and how its effects extend across relevant interactions. A reusable correction specifies a prerequisite, role, or mechanism that applies across player states within its declared scope, while changes tied to entities, case identifiers, or the wording of triggering interactions are tested with rephrased requests and condition-derived counterexamples.

Some game mechanics legitimately apply only to particular characters or circumstances. Such corrections can be registered as scoped exceptions when their evidence and applicability conditions receive expert confirmation, and the release record preserves the accepted scope and supporting evidence. Other case-specific changes remain quarantined.

\section{Evaluation Protocol}
\label{app:evaluation}

\subsection{Sampling}
\label{app:sampling}

We define coaching turns as responses to players' strategic or knowledge requests. Standalone recommendations, casual conversation, refusals, and safety-filtered interactions are excluded. From each consecutive two-day deployment period we sample 82 sessions at random and keep all coaching turns within them, which yields 1,122 turns split as 397, 365, and 360 across the three periods. Earlier turns stay available as dialogue history when later responses are replayed. Revision decisions and release tolerances were fixed before the sealed period was evaluated.

\subsection{Expert annotation}
\label{app:annotation}

Game-domain experts annotate interaction histories using player state, retrieved strategy packages, rule outputs, tool results, and versioned game evidence, with system versions and deployment periods hidden. Labels cover correctness, instructiveness, responsible components, acceptable repair targets, and the validity and target of player feedback. Substantive errors include claims contradicted by game mechanics or player resources and omitted interactions that change a recommended decision, while evasive or generic responses are labeled non-instructive. Repair ownership follows Table~\ref{tab:attribution-stages} and may name several targets when more than one change could resolve a failure.

Annotation quality is measured by Cohen's $\kappa$ of 0.69 for correctness, 0.84 for instructiveness, 0.64 for repair ownership, and 0.77 for the validity of player doubt.

\subsection{Judge protocol}
\label{app:judge}

Evaluation combines response-level judgments, executable checks, and recorded process evidence, following agent evaluation protocols for task completion \citep{liu2023agentbench,zhou2024webarena,mialon2024gaia} and intermediate progress \citep{ma2024agentboard}. A fixed self-hosted DeepSeek-V4-Flash judge \citep{xu2026deepseek} applies structured criteria \citep{liu2023geval} to responses that are shuffled without configuration labels. Its rubric was selected on a calibration set from the second period and held fixed for later evaluation and for all system comparisons.

Reference evidence includes the recorded reply context, tool outputs, knowledge snapshots, and game data from the relevant period. When the recorded context conflicts with game-data tables, the protocol gives precedence to the recorded context because it preserves the version-specific information supplied during that interaction. Replayed responses are judged against that context and the target period's rules. Claims inconsistent with the reference evidence are contradicted, including incorrect counter relations and omitted skill interactions that change the conclusion, and outcomes that the available calculations cannot establish are labeled unknown.

Correctness requires no contradicted claims and satisfaction of four executable constraints. Recommended characters must be available in the player's lineup, instructed skills must belong to the selected characters, stated counter relations must agree with the reference table, and resource-dependent abilities must be supported by the player's resources. Instructiveness is assessed separately, and evasive answers, generic guidance, and unnecessary clarification receive no credit on the applicable quality dimensions. A response passes when it is correct and instructive and averages at least 1.5 across applicable relevance, explanation, and actionability scores. Requests that ask for a numerical calculation are not penalized for limited strategic explanation when the calculation answers the request.

On the held-out evaluation set, the judge agrees with expert correctness labels on 89.0\% of turns ($\kappa=0.48$) and with instructiveness labels on 86.4\% ($\kappa=0.56$). Incorrect-response rates are 11.7\% for the judge and 12.3\% for experts, and error-detection precision and recall are 55.6\% and 52.6\%. Adding executable checks yields 85.7\% correctness agreement ($\kappa=0.42$), and the knowledge-form comparisons retain their direction under judge-only correctness labels.

Bias checks follow \citet{zheng2023llmjudge}. Placing reference evidence before the response leaves correctness agreement at 89.0\% ($p=1.0$), and false alarms are more frequent for longer responses, reaching 15.9\% in the length analysis, while shorter responses account for more missed errors. Each response is generated twice. Outcomes that are consistent across repetitions are treated as stable for paired tests, and cases with low-confidence judgments or missing evidence remain in the denominator with unresolved status.

\subsection{Knowledge forms and condition variants}
\label{app:forms}

The knowledge-form comparison uses 197 strategic turns from 69 sessions in the sealed period, with shared request, player state, tool outputs, reply-stage rules, and response model. \emph{Atomic facts} supplies relevant versioned character attributes, skills, and counter relations. The \emph{unstructured guide} rewrites packages, rule outputs, and instructional content as continuous prose, preserving facts, values, and recommendations while removing explicit condition markers and structural headings. Manual review of a sample of rewrites finds no missing facts or conditions, and the few that omit a presentation instruction do not change the comparisons when excluded. The $-$\,\emph{cond.\ rules} ablation removes conditional rule outputs and keeps the remaining package content. Knowledge tokens count reply-prompt tokens beyond the shared context.

For the applicability evaluation we select 36 turns whose strategies depend on a lineup-wide resource and construct three matched player states per turn, in which the requirement is satisfied, violated, or unavailable for evaluation. The static guide keeps its original text while the package recomputes the applicable rule outputs. Each configuration generates two responses per variant, and a separate judgment with the same underlying model assesses violations against condition-dependent behavior specified before evaluation.

\subsection{Replay and revision units}
\label{app:replay}

Replay reconstructs a target reply from its recorded request, dialogue history, player state, planning decision, and tool outputs. The configuration under test supplies knowledge content and reply-stage rules, knowledge unavailable in that version is excluded, and logged tool explanations and derived game-data tables remain fixed. Only the target reply is regenerated, with earlier responses preserved as recorded, and a fidelity check verifies that logged rules and retrieved knowledge appear in the snapshot of the corresponding deployed version.

Changes from the two deployed rounds are grouped into 15 revision units by the strategic sub-intent packages they affect. Shared reply instructions and basic-knowledge changes form common units that cover the relevant coaching turns, and development cases are earlier failures attributed to a unit or judged incorrect by experts. Each unit is assessed retrospectively on the sealed set.

Release eligibility requires a positive 95\% lower confidence bound for in-scope correctness gain. Recovery among later exposed failures must reach at least half the recovery observed on development failures, changes on unexposed cases must stay within unchanged-system rerun variation, and neither out-of-scope correctness nor answer coverage may decline by more than three percentage points. Dependent knowledge and execution changes are evaluated as a combined candidate.

For the planted controls we select three failure classes from the second period, each with at least two triggering cases, namely recommending an unavailable character, misstating a counter relation, and asserting a turn-order advantage without supporting evidence. Each class receives an entity-specific correction and a mechanism-based correction of similar length, and the six candidates are inserted separately into the reply-stage rules of the third version. Expected specificity labels were fixed before evaluation, the generality evaluator applies a fixed rubric with a majority decision over three runs, and experts label all 21 planted and deployed units without knowing which candidates were planted.

\subsection{Statistical analysis}
\label{app:statistics}

We estimate 95\% confidence intervals from 10,000 session-level bootstrap resamples that preserve all turns and repeated generations within a sampled session. Exact McNemar tests use paired turns with stable outcomes across both generations. Reported gains are also compared with unchanged-system rerun differences, and Holm correction is applied to prespecified families of multiple comparisons. Reply generation on the sealed period has a median of 2.07\,s and a 95th percentile of 5.48\,s, and the measurement covers generation rather than the complete online workflow. Reproduction materials include model identifiers, prompts, evaluation rubrics, and revision-unit definitions, subject to restrictions on releasing player interaction records.

\section{Additional Results}
\label{app:results}

\subsection{Failure diagnosis}
\label{app:diagnosis}

The diagnosis set contains 130 player-doubt turns from 39 sessions in the first period, with 48 doubts adjudicated by experts, of which 36 are founded and 12 unfounded. \emph{Doubt as failure} treats the immediately preceding turn as erroneous. \emph{Direct feedback} judges the request, response, and subsequent player utterance. \emph{Category routing} predicts a failure category and maps it to a component. \emph{Who\&When} \citep{zhang2025whowhen} applies single-pass trace attribution across planning, tool-call, tool-result, and reply stages. \systemname{} verifies claims and locates the targeted turn, and its ablation considers only the preceding response.

Category routing reaches 88.0\% precision and 61.1\% recall, against 86.4\% and 52.8\% for \systemname{}. The latter rejects 9 of 12 unfounded doubts and leaves 36.9\% of flagged cases unresolved when trace evidence is insufficient. Target-turn localization identifies 26.7\% of delayed-feedback failures, which the preceding-turn ablation misses entirely. On separately evaluated repair attribution, ownership macro-F1 is 18.8\% and repair-target Hit@1 is 36.7\%, and reply execution accounts for 28 of the 30 expert-attributed failures.

\subsection{Condition variants}
\label{app:variants}

Table~\ref{tab:variants} reports violations separately for false, unknown, and true conditions. Unknown conditions produce the largest difference between the static guide and the package, at 26.4\% against 15.3\%. Removing conditional rules yields a 19.4\% violation rate under false conditions. Under true conditions every configuration keeps the strategy in more than 93\% of responses. Atomic facts and the package show the same violation rates in this controlled test even though their correctness on recorded strategic turns differs substantially, which is the separation that the knowledge-form comparison of the main paper quantifies.

\begin{table}[t]
\centering
\footnotesize
\caption{Strategy-condition violation rates (\%) by condition outcome on 108 matched player-state variants.}
\label{tab:variants}
\setlength{\tabcolsep}{5pt}
\begin{tabular}{@{}lccc@{}}
\toprule
Knowledge & False & Unknown & True \\
\midrule
Atomic facts & 5.6 & 15.3 & 4.2 \\
Unstructured guide & 9.7 & 26.4 & 5.6 \\
Strategy package & 5.6 & 15.3 & 4.2 \\
\quad $-$ cond.\ rules & 19.4 & 23.6 & 6.9 \\
\bottomrule
\end{tabular}
\end{table}

\subsection{Deployed-version comparisons}
\label{app:version-replay}

Replaying the three versions on the same 360 sealed turns, with two generations per configuration, gives a paired correctness difference of $+0.6$ points from the first to the third version, with a 95\% CI of $[-2.2, 3.4]$ and exact McNemar $p=1.0$. On the 72 turns exposed to second-round changes, the third version exceeds the second by 2.8 points with a 95\% CI of $[-4.2, 10.3]$. Neither comparison establishes a significant improvement.

Logged-response correctness across the three periods is 83.1\%, 84.4\%, and 84.4\%, and an ordered-period analysis gives an odds ratio of 1.05 ($p=0.66$). The pass rate rises from 51.4\% to 55.8\% ($p=0.30$), while the non-instructive rate falls from 16.6\% to 12.8\% ($p=0.22$). In a blinded comparison of 60 turns, expert preferences for the third version over the first are 44.0\% for correctness, 36.3\% for instructional quality, and 38.8\% overall. These observational and expert comparisons likewise do not establish a significant improvement across the deployed revisions.

\subsection{Release checks and generality}
\label{app:release-results}

Retrospective assessment finds that one of 15 deployed units satisfies the eligibility criteria. Thirteen units lack sufficient gain beyond unchanged-system variation. The remaining unit improves in-scope correctness by 25.0 points while reducing out-of-scope correctness by 20.8 points, which exceeds the regression tolerance and shows why both affected and unaffected traffic are measured.

An unchanged rerun of the first version recovers 38.4\% of previously failing turns with a paired difference of $+0.6$ points. The planted entity-specific and mechanism-based controls recover 39.1\% and 40.7\%, while their mean paired in-scope differences are $-0.6$ and $+0.7$ points, so all six fail independent gain requirements. The generality check classifies all three entity-specific controls as case-specific and all three mechanism-based controls as general. Agreement with independent expert labels over all 21 planted and deployed units is 85.7\% ($\kappa=0.56$), and expert classifications match five of the six prespecified planted labels.

\subsection{Knowledge-execution effects}
\label{app:dependency-effects}

A factorial replay evaluates first-round knowledge revisions alone, execution-rule revisions alone, and their combination on the same frozen inputs. Among 26 turns exposed to jointly developed changes, knowledge-only updates yield a $+7.7$-point difference. The interaction term is $-9.6$ points with a 95\% CI of $[-17.9, -2.0]$, so the combined effect falls below the sum of the separately measured effects on this subset. The result supports joint assessment of dependent edits while keeping their contributions separately measurable.

\end{document}